\documentclass[lettersize,journal,twoside]{IEEEtran}
\usepackage{amsmath,amsfonts}
\usepackage{algorithmic}
\usepackage{algorithm}
\usepackage{array}
\usepackage[caption=false,font=normalsize,labelfont=sf,textfont=sf]{subfig}
\usepackage{textcomp}
\usepackage{stfloats}
\usepackage{url}
\usepackage{verbatim}
\usepackage{graphicx}
\usepackage{pifont}
\usepackage{cite}
\usepackage{xcolor}
\usepackage[table]{xcolor}
\usepackage{multirow}
\usepackage{siunitx}
\usepackage{tikz}
\usepackage{booktabs}
\begin{document}

\newcommand{\yuxing}[1]{\textcolor{blue}{Yuxing: #1}}
\newcommand{\bowen}[1]{\textcolor{red}{Bowen: #1}}
\newcommand{\vspaceafter}{\vspace{-3pt}}

\definecolor{mygreen}{RGB}{79,173,91}
\newcommand{\greendot}{%
  \tikz[baseline=-0.6ex]{\fill[mygreen] (0,0) circle (0.6ex);}%
}

\definecolor{myred}{RGB}{234,51,35}
\newcommand{\reddot}{%
  \tikz[baseline=-0.6ex]{\fill[myred] (0,0) circle (0.6ex);}%
}

\newcommand{\cmark}{\textcolor{green}{\checkmark}}
\newcommand{\xmark}{\textcolor{red}{\ding{55}}}
\newcommand{\yellowcircle}{%
  \tikz[baseline=-0.5ex]{\draw[yellow!80!brown, line width=1.8pt] (0,0) circle (0.8ex);}%
}

\title{FoldNet: Learning Generalizable Closed-Loop Policy for Garment Folding via Keypoint-Driven Asset and Demonstration Synthesis}

\author{Yuxing Chen, Bowen Xiao, and He Wang 

\thanks{Manuscript received: January 9, 2026; Revised January 17, 2026; Accepted: January 7, 2026. This paper was recommended for publication by Editor Aleksandra Faust upon evaluation of the Associate Editor and Reviewers comments. This work was supported by Galbot.\textit{(Yuxing Chen and Bowen Xiao contributed equally to this work.)(Corresponding author: He Wang.)}}
\thanks{The authors are with CFCS, School of Computer Science, Peking University, Beijing 100051, China, and also with Galbot, Beijing 100010, China.}
\thanks{E-mail: yuxingc\_20@stu.pku.edu.cn; xiaobowenbowie@stu.pku.edu.cn;\newline hewang@pku.edu.cn}
\thanks{Digital Object Identifier (DOI): see top of this page.}

}

\maketitle

\markboth{IEEE Robotics and Automation Letters. Preprint Version. January, 2026}%
{Chen \MakeLowercase{\textit{et al.}}: FoldNet: Learning Generalizable Closed-Loop Policy for Garment Folding}


\begin{abstract}
Due to the deformability of garments, generating a large amount of diverse and high-quality data for robotic garment manipulation tasks is highly challenging. In this paper, we present \textit{FoldNet}, a synthetic garment dataset that includes assets for four categories of clothing as well as high-quality closed-loop folding demonstrations. We begin by constructing geometric garment templates based on keypoints and applying generative models to generate realistic texture patterns. Leveraging these garment assets, we generate folding demonstrations in simulation and train folding policies via closed-loop imitation learning. To improve robustness, we introduce \textit{KG-DAgger}, a keypoint-based strategy for generating recovery demonstrations after failures. KG-DAgger significantly improves the quality of generated demonstrations and the model performance, boosting the real-world success rate by 25\%. After training with 15K trajectories (about 2M image-action pairs), the model achieves a 75\% success rate in the real world. Experiments in both simulation and real-world settings validate the effectiveness of our proposed dataset.
\end{abstract}



\section{Introduction}
\label{sec:intro}

\begin{figure*}[t]
    \centering
    \includegraphics[width=0.9\linewidth]{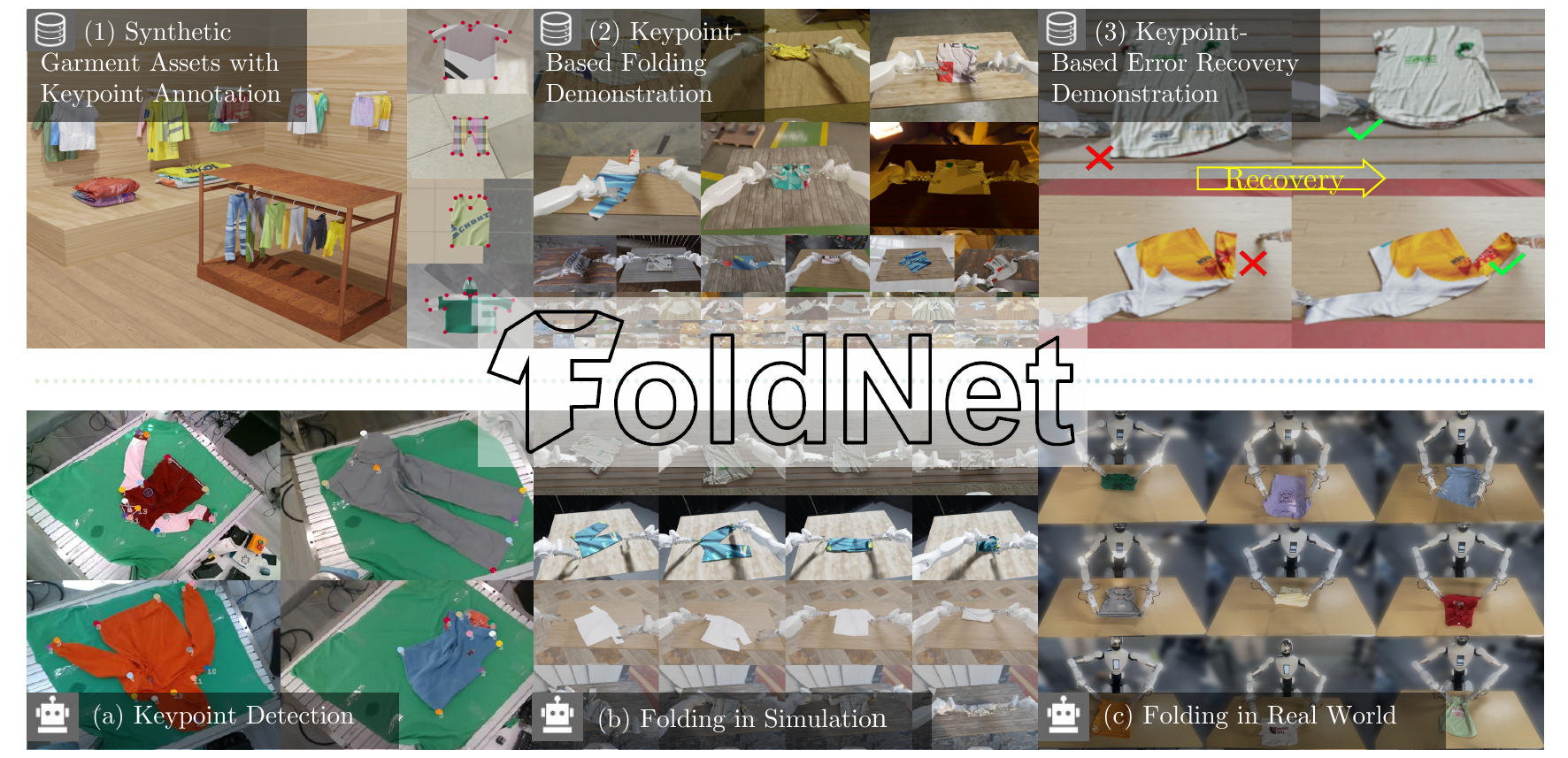}
    \caption{\textbf{FoldNet} is a dataset designed for robotic garment manipulation. It provides (1) a large collection of synthetic garment assets with keypoint annotations, (2) high-quality folding demonstrations, and (3) keypoint-based error recovery demonstrations. Leveraging these assets, FoldNet supports a wide range of downstream tasks, including (a) keypoint detection, (b) folding in simulation, and (c) folding in the real world.}
    \label{fig:teaser}
    \vspaceafter
\end{figure*}

\IEEEPARstart{G}{arment} manipulation has been widely studied in robotics~\cite{longhini2024unfoldingliteraturereviewrobotic}. Due to the deformable nature of garments, such tasks remain highly challenging. In recent years, data-driven learning approaches have made significant progress, with imitation learning~\cite{zare2024surveyimitation} gradually emerging as the main paradigm for the acquisition of various robotic skills. Some prior works~\cite{pi0,liu2024rdt} have demonstrated strong garment manipulation capabilities using imitation learning. However, enabling the learned policy to generalize to unseen environments and objects remains hindered by the scarcity of large-scale, diverse, high-quality demonstration data.

Learning from synthetic data has become an efficient approach for robot learning~\cite{canberk2022clothfunnels,wang2023policydressalllearning}. Many datasets~\cite{bertiche2020cloth3d,zhou2023clothesnetinformationrich3dgarment,lips2024learningkeypointsroboticcloth} and simulation environments~\cite{lu2024garmentlab,li2018learning} are now available to generate garment manipulation data. In simulation, it is possible to flexibly modify both the environment and the properties of the garments, allowing stronger generalization capabilities. However, improving the quality of synthetic data remains a key challenge. Current methods face two main limitations:

\textbf{Limited garment assets and lack of detailed annotations.} Existing datasets often contain only a small number of garment meshes and lack rich annotations. This limited data availability imposes an upper bound on generalization performance. Moreover, the lack of detailed annotations requires researchers to put in extra effort to generate high-quality demonstration data in simulation.

\textbf{Limited handling of error recovery.} Garment manipulation tasks are long-horizon tasks that involve complex deformable object dynamics. Compared to previously dominant open-loop approaches for garment manipulation~\cite{canberk2022clothfunnels,xue2023unifoldingsampleefficientscalablegeneralizable}, closed-loop control offers the potential to retry after failures. However, if training data only contains perfect demonstrations, small errors at each step can accumulate and potentially cause the garment to enter previously unseen states, often resulting in task failure. This poses a significant challenge for learning robust policies.

To address the scarcity of garment assets, we propose a novel framework for generating garment assets. For each category of garment, we design a template whose geometry is controlled by a set of keypoints. We then apply generative models to synthesize texture maps for garments. This approach enables scalable garment mesh generation, and each mesh is accompanied by automatically generated semantic keypoint annotations for subsequent demonstration generation and policy learning.

To handle out-of-distribution states, we introduce Keypoint-Gated DAgger (KG-DAgger). After training the initial policy network, we run the policy and use the previously automatically annotated keypoints to detect potential failure cases. When a failure is detected, a keypoint-based strategy is invoked to perform a correction. The corrected trajectories are then added to the dataset for further policy training. The final model is end-to-end: given the current observation, the model directly outputs the action sequence without requiring any additional hyperparameters.

In summary, this work makes the following two key contributions:

\begin{enumerate}
\item We propose a garment mesh generation framework that can automatically generate highly diverse garment meshes with annotated keypoints.
\item We introduce KG-DAgger, improving the data quality and boosting the success rate of closed-loop folding in the real world from 50\% to 75\%. 
\end{enumerate}


\section{Related Works}
\label{sec:related}

\subsection{Garment Manipulation}
Garment manipulation is a widely studied task in robotics~\cite{longhini2024unfoldingliteraturereviewrobotic}. The main challenges arise from the deformable nature of garments and their complex dynamics. Various approaches have been explored, including imitation learning~\cite{avigal2022speedfoldinglearningefficientbimanual}, reinforcement learning~\cite{wang2023policydressalllearning}, and model-based methods~\cite{tian2025diffusiondynamicsmodelsgenerative,jiang2025phystwinphysicsinformedreconstructionsimulation}. 

From a task perspective, many works focus on garment folding~\cite{xue2023unifoldingsampleefficientscalablegeneralizable,lips2024learningkeypointsroboticcloth,chen2025metafoldlanguageguidedmulticategorygarment,avigal2022speedfoldinglearningefficientbimanual} and unfolding~\cite{canberk2022clothfunnels,chen2024deformpam}. However, most approaches for solving folding or unfolding tasks rely on modular perception and control pipelines, which exhibit several limitations. Many rely on object point clouds~\cite{xue2023unifoldingsampleefficientscalablegeneralizable,chen2025metafoldlanguageguidedmulticategorygarment} and therefore depend on accurate camera calibration and depth information to achieve robust grasping, making recovery from failed grasps challenging. In addition, they often require numerous hand-designed hyperparameters — such as lift heights during folding — which are difficult to generalize across garments of varying sizes. 

In recent years, closed-loop policies~\cite{pi0,liu2024rdt,kim2024openvlaopensourcevisionlanguageactionmodel,octomodelteam2024octoopensourcegeneralistrobot} trained on large-scale real-world data have demonstrated strong capabilities in garment manipulation. However, collecting these datasets requires a large amount of human labor. Moreover, real-world data cannot offer the strong generalization that synthetic data can provide. In this work, we investigate how to generate high-quality synthetic demonstrations of garment manipulation for training closed-loop models. 

\subsection{Synthetic Garment Assets}
Compared with rigid-body mesh assets~\cite{objaverseXL}, garment assets that can be physically simulated place much higher demands on mesh quality. Existing garment mesh datasets are typically designed manually by artists~\cite{zhou2023clothesnetinformationrich3dgarment} or generated based on predefined templates~\cite{lips2024learningkeypointsroboticcloth,bertiche2020cloth3d}. Though template-based methods allow for large-scale mesh generation at substantially lower cost compared to manual design, they face significant challenges when applying realistic texture to the mesh. Previous template-based methods either directly apply existing texture libraries to garment meshes~\cite{lips2024learningkeypointsroboticcloth}, or use generative models to synthesize textures~\cite{youwang2024paintit,wen2024foundationposeunified6dpose}. However, the textures generated by the first method differ significantly from those of real garments, whereas the second method perform poorly when applied to layered garment meshes. In this work, we adopt a template-based method to generate garment geometry and introduce a pipeline that facilitates generative models in producing scalable and realistic textures.

\subsection{Imitation Learning}
Imitation learning~\cite{hejna2024remixoptimizingdatamixtures,myers2024policyadaptationlanguageoptimization} has received increasing attention from the research community. A key challenge lies in collecting high-quality demonstration data. Recent studies have shown that enabling models to recover from errors leads to better performance than naive imitation learning, making it a topic of great interest~\cite{kelly2019hgdaggerinteractiveimitationlearning,luo2024hilserl}. Our method performs imitation learning in simulation by distilling a keypoint-based policy into a vision-based model, while improving robustness by generating demonstrations that incorporate recovery from failures.


\section{Garment Mesh Synthesis}
\label{sec:garment_synthesis}

\begin{table}[t]
    \centering
    \caption{\textbf{Comparison with other synthetic datasets.} The table columns indicate the number of garment meshes, number of garment categories, inclusion of RGB textures, multi-layer meshes (front layer and back layer), semantic keypoints, and mesh resolution. \textsuperscript{1}The numbers here indicate that the garments can be generated, with the quantity representing the number of meshes that can be generated in one day on a single RTX 3090. \textsuperscript{2}Including the time required for rendering. \textsuperscript{3}Although RGB data is included, it does not maintain consistency with the clothing geometry. \textsuperscript{4}The resolution is adjustable.}
    \label{tab:comparison_dataset}
    \begin{tabular}{|c|c|c|c|c|c|c|}
        \hline
        Dataset & \#M & \#C & RGB & ML & SK & Res\\
        \hline
        ClothesNet~\cite{zhou2023clothesnetinformationrich3dgarment} & 3.1K & 11 & \cmark & \cmark & \xmark & \SI{1}{\centi\metre}\\
        \hline
        Cloth3D~\cite{bertiche2020cloth3d} & 11.3K & 4 & \xmark & \cmark & \xmark & \SI{1}{\centi\metre}\\
        \hline
        aRTF~\cite{lips2024learningkeypointsroboticcloth} & 10K/D \textsuperscript{1,2} & 3 & \yellowcircle \textsuperscript{3} & \xmark & \cmark & Adj.\textsuperscript{4} \\
        \hline
        Ours & 2K/D \textsuperscript{1} & 4 & \cmark & \cmark & \cmark & Adj.\textsuperscript{4} \\
        \hline
    \end{tabular}
    \vspaceafter
\end{table}

\begin{figure}[h]
    \centering
    \includegraphics[width=1\linewidth]{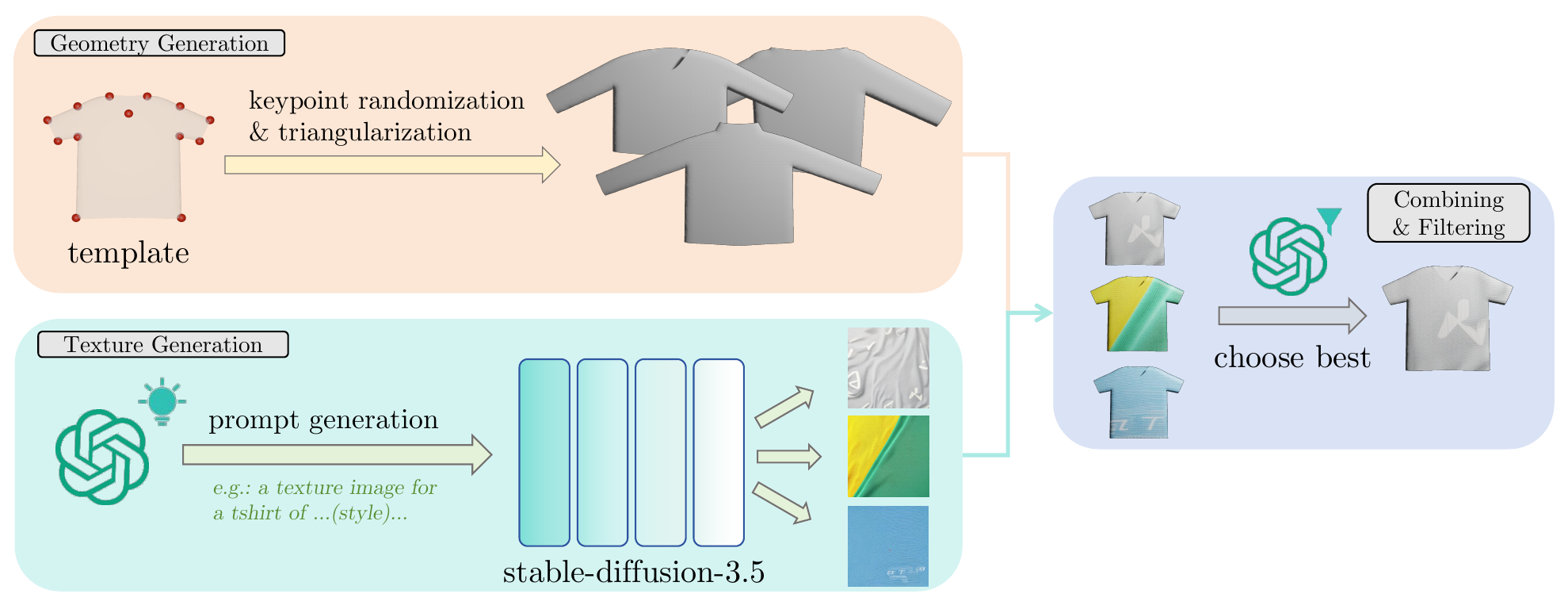}
    \caption{\textbf{Pipeline for garment mesh synthesis.} By performing \textit{geometry generation}, \textit{texture generation}, \textit{combining-and-filtering}, we can synthesize scalable, high-quality garment meshes.}
    \label{fig:synthetic_data_pipeline}
    \vspaceafter
\end{figure}

Our method begins with synthesizing high-quality garment meshes. These meshes need to be suitable for physical simulation and rendering. Our pipeline for garment generation is shown in Figure~\ref{fig:synthetic_data_pipeline}. The main steps include: (1) creating the geometry of the garment, (2) generating the texture of the garment, (3) combining the geometry and texture, and then filtering. Detailed descriptions of these stages are provided in~\ref{ssec:geometry_generation},~\ref{ssec:texture_generation} and~\ref{ssec:combining_and_filtering}. To show the advantages of our approach, we compare the resulting asset dataset with several existing datasets in Table~\ref{tab:comparison_dataset}.

\subsection{Geometry Generation}
\label{ssec:geometry_generation}
We use a template-based approach to generate garment geometry of four types of garments — t-shirt (including long-sleeved and short-sleeved), vest (sleeveless), hoodie, and trousers. For each type of garment, the template is constructed by manually specifying a set of semantic keypoints, i.e., 2D positions $(x, y)$, that capture the structural characteristics of the garment. These keypoints serve a dual role: they identify semantically meaningful manipulation points on the garment and implicitly define its shape. Once the keypoint positions are determined, we connect them along the border using Bezier curves and perform triangulation within the xy-plane. Then, we heuristically define the z-coordinates and UV coordinates for the mesh vertices. During this process, keypoints are automatically annotated on the generated triangular mesh by saving the keypoint indices. With this generation method, we can generate a large variety of garment shapes with high efficiency by simply randomizing the positions of the keypoints. 

\subsection{Texture Generation}
\label{ssec:texture_generation}
To automatically generate garment textures, we use pre-trained generative models. First, for each type of garment, we use a large language model~\cite{openai2024gpt4ocard} to generate a description of the texture. Then, we use this description as a prompt for a Text2Image model~\cite{podell2023sdxlimprovinglatentdiffusion}. Repetition of this process multiple times can quickly generate a large number of texture images.

\begin{figure}[h]
    \centering
    \includegraphics[width=1\linewidth]{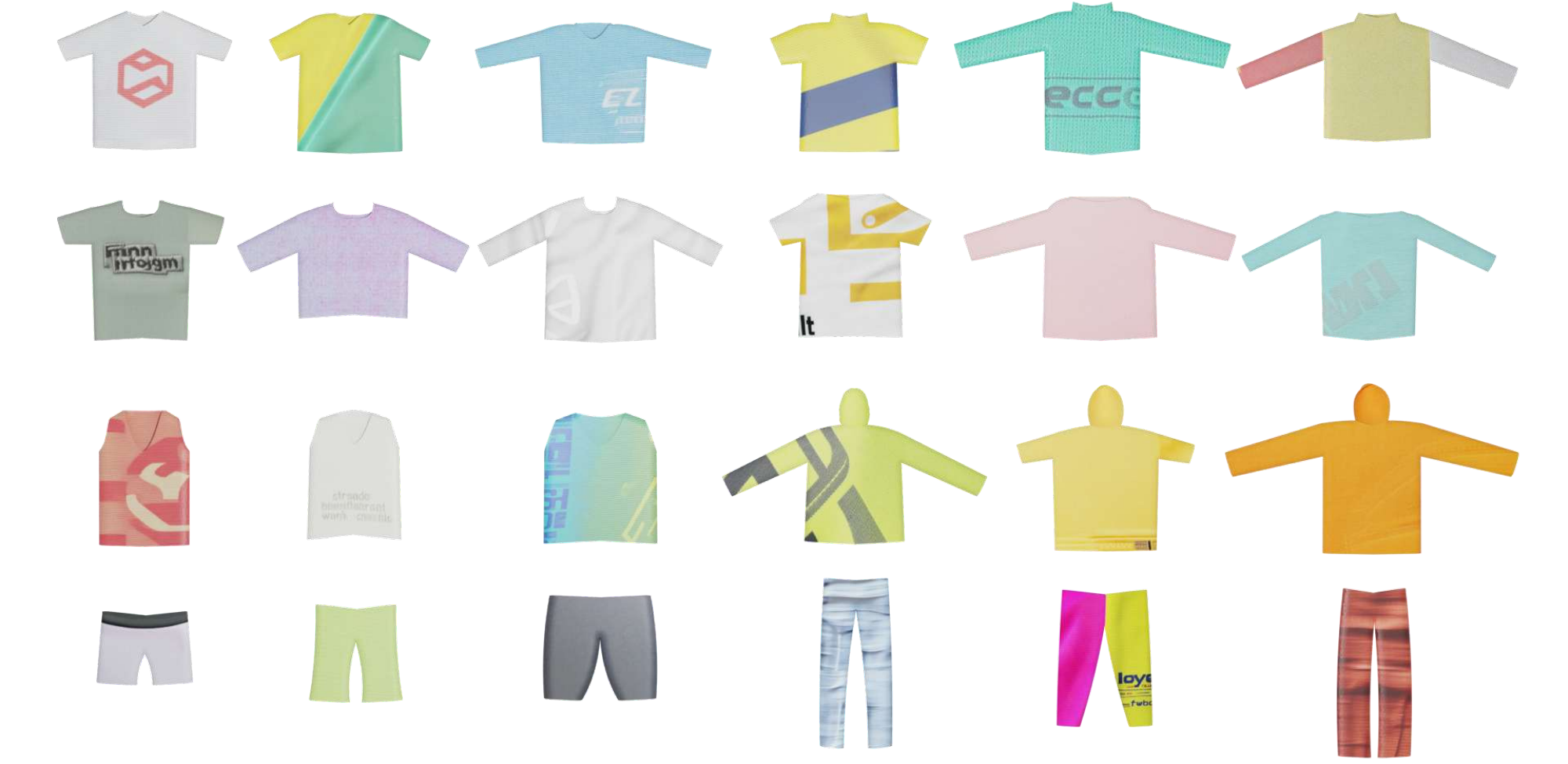}
    \caption{\textbf{Synthetic garment meshes.} These static garment meshes can be used for subsequent physics simulation and policy learning.}
    \label{fig:syn_assets}
    \vspaceafter
\end{figure}

\subsection{Combining and Filtering}
\label{ssec:combining_and_filtering}To enhance the consistency between texture images and garment meshes, we introduce an additional filtering step. For each garment mesh with only geometry, we combine it with different texture images and render the results. A vision language model~\cite{openai2024gpt4ocard} is then used to automatically select the most suitable texture as the final texture for that mesh. We present several examples of the final generated garments, as shown in Figure~\ref{fig:syn_assets}.


\section{Demonstration Generation}
\label{sec:demonstration}

Using the generated garment assets, we design keypoint-based policies to automatically collect demonstrations in simulation. A vision-action model $M_0$ is then trained on these demonstrations via imitation learning. To improve data efficiency and policy robustness, we introduce KG-DAgger, a variant of DAgger. In KG-DAgger, at the $i$-th iteration, we use the current model $M_i$ to generate new trajectories. During this process, a keypoint-based error recovery strategy is employed. These newly generated trajectories teach the model how to recover from errors—particularly the types of errors to which the model is most prone—thereby continually improving the model's performance. In the final deployment, the model is trained on the entire set of trajectories. The final policy is an \textit{end-to-end} system: it does not require explicit keypoint detection or error detection, as these capabilities are learned implicitly by the model.

\begin{figure}[h]
    \centering
    \includegraphics[width=1.0\linewidth]{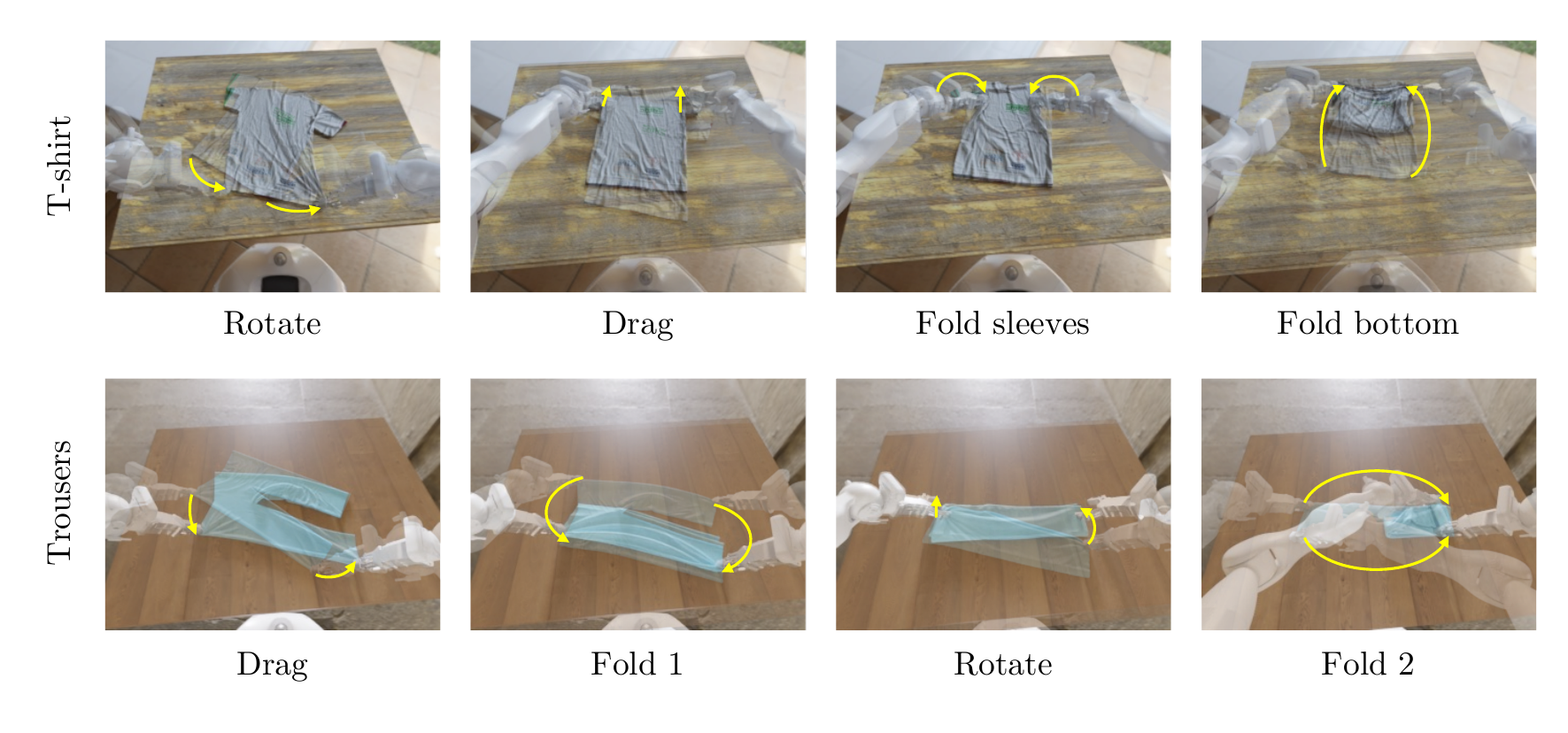}
    \caption{\textbf{Keypoint-based demonstration generation.} The entire folding process can generally be divided into several stages, which are then executed sequentially during the demonstration generation process.}
    \label{fig:keypoint_based_policy}
    \vspaceafter
\end{figure}

\begin{figure*}[t]
    \centering
    \begin{minipage}{0.68\textwidth}
        \centering
        \includegraphics[width=1.0\linewidth]{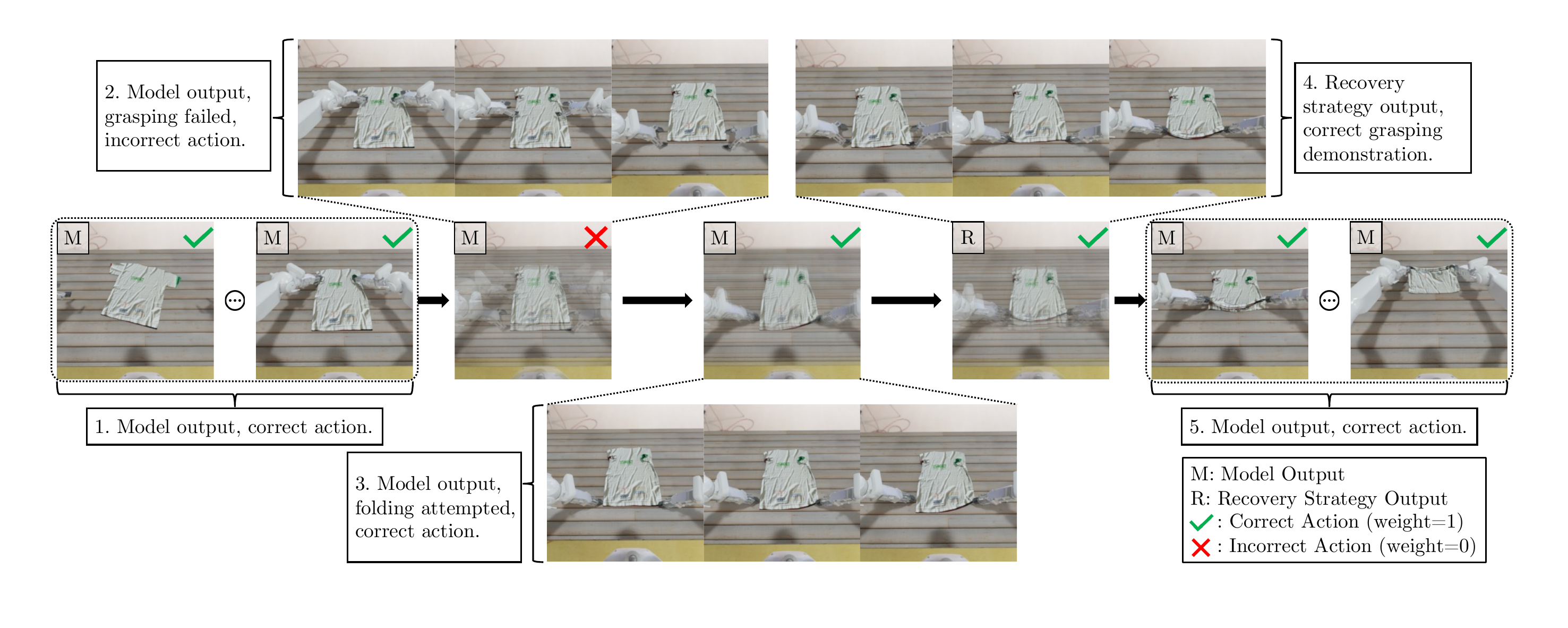}
        \vspace{-8mm}
    \end{minipage}%
    \hfill
    \begin{minipage}{0.3\textwidth}
        \centering
        \includegraphics[width=1.0\linewidth]{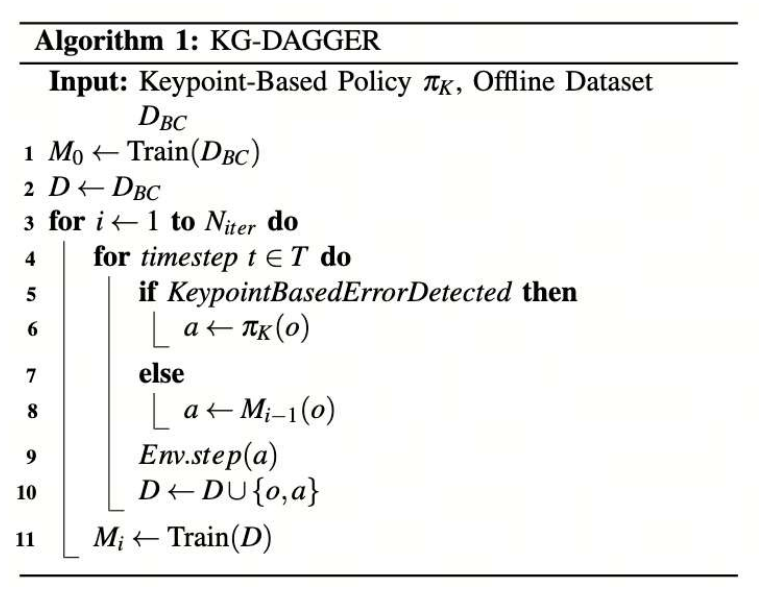}
    \end{minipage}
    \caption{\textbf{Keypoint-based recovery strategy.} The figure on the left illustrates an example of the recovery strategy. The recovery data are incorporated into the dataset and used jointly to train the end-to-end vision-action model. With these data, the model can learn how to retry when a grasp attempt fails. The figure on the right shows the pseudocode of KG-DAgger.}
    \label{fig:keypoint_based_recovery}
\end{figure*}

\subsection{Keypoint-Based Demonstration Generation}
\label{ssec:keypoint_policy}

For each garment category, we design a simple yet effective keypoint-based policy to fold the garment in a predefined manner. For example, in the case of a t-shirt, one possible folding strategy involves first rotating the garment, then dragging it, folding both sleeves inward, and finally folding the bottom of the shirt. Example demonstrations are shown in Figure~\ref{fig:keypoint_based_policy}. At each stage, the initial grasping points and target placement positions for the two grippers are derived from ground-truth keypoint locations, while intermediate positions are obtained through interpolation. Different folding strategies can be generated by modifying the initial and target positions at each stage. Owing to the keypoint annotations in our assets, this policy is unified across garments of the same category, independent of shape variations.

\subsection{Error-Recovery Demonstration Generation}
\label{ssec:keypoint_recovery}

KG-DAgger is similar to HG-DAgger~\cite{kelly2019hgdaggerinteractiveimitationlearning}: during model inference, we use a keypoint-based strategy to detect grasp failures — a step that is performed by a human expert in HG-DAgger. This monitoring process leverages the keypoints of the garment at each time step, along with the gripper state. Figure~\ref{fig:keypoint_based_recovery} illustrates in detail how the keypoint-based error recovery strategy is implemented. The entire recovery process is divided into five stages.

\begin{itemize}
    \item In Stage 1, the model outputs correct actions, and the recovery strategy does not need to intervene.
    \item In Stage 2, the model outputs incorrect actions, failing to move the gripper to the correct position and thus causing a grasp failure. This stage covers the interval from the previous release of the gripper until the failed attempt when the gripper closes. These actions should not be encouraged; during training, we assign them a weight of zero.
    \item In Stage 3, the gripper fails to grasp the garment due to its incorrect position in the previous step. However, the model should still continue attempting: only after moving the gripper and observing that the garment does not move can the failure be detected. Therefore, the actions in this stage are still correct.
    \item In Stage 4, the recovery strategy takes over and retries the grasp. All actions in this stage are correct.
    \item In Stage 5, the error has been resolved, and the model resumes generating and executing actions.
\end{itemize}

By incorporating these error-recovery trajectories into the dataset for training, the model can learn how to retry grasping after a failure.

This KG-DAgger process is used only during the training phase and only in simulation. During testing in simulation and in the real world, we directly use the outputs of the vision–action model, without requiring KG-DAgger or keypoint detection.

\subsection{Model Training}
We choose diffusion policy~\cite{chi2023diffusionpolicy} as our vision-action model for its compact size and good performance in modeling multi-modal behaviors and producing coherent action sequences. It is also possible to use other vision-action models, as our demonstration data do not require a specific model. 

We retain only successful episodes and filter out failed ones. Here, an episode refers to the entire trajectory starting from the initial state and finally resulting in the garment being fully folded. At the end of each episode, we pad several no-op actions to indicate termination. The training loss is a modified version of the original diffusion loss~\cite{chi2023diffusionpolicy}: $L_\theta=\sum_{i=1}^{T_a}m_i*||\varepsilon^k_i-\varepsilon_\theta(O_t,A_t^0+\varepsilon^k,k)_i||^2$. For the $i$-th action in an action chunk, we multiply the loss by a coefficient $m_i$. If a zero-weight action (due to a grasp failure) appears in the action chunk, then $m_i$ for that action and all subsequent actions is set to 0; otherwise, $m_i$ is 1. When all $m_i = 1$, the loss reduces to the original diffusion policy loss.


\section{Experiments}
\label{sec:exp}

We design two tasks to validate the effectiveness of our method: keypoint detection and garment folding. The keypoint detection task is easier to benchmark and illustrates how closely the generated garment meshes resemble real-world garments. The garment folding task is more comprehensive and enables the evaluation of the quality of the generated demonstration data.

\subsection{Keypoint Detection}

\begin{table*}[t]
\centering
\caption{\textbf{Quantitative results of keypoint detection on real images.} The figure illustrates the performance of models trained using different garment mesh synthesis methods. In the \textit{Average} row, we highlight the top two values in bold.}
\label{tab:keypoint_detection_performance}
\begin{tabular}{lcccccccc}
    \toprule
    & \multicolumn{4}{c}{\textbf{mAP$_{4,8}$ (↑)}} & \multicolumn{4}{c}{\textbf{AKD (↓)}} \\
    \cmidrule(lr){2-5} \cmidrule(lr){6-9}
    \textbf{Category} & Ours & w/o filter & aRTF & Paint-it & Ours & w/o filter & aRTF & Paint-it \\
    \midrule
    T-Shirt  & 59.0 & 50.5 & 42.2 & 47.2 & 10.3 & 9.30 & 11.3 & 14.2 \\
    Trousers & 51.7 & 57.0 & 47.4 & 47.8 & 16.9 & 16.4 & 14.1 & 32.0 \\
    Vest     & 42.5 & 43.5 & 26.0 & 37.3 & 20.0 & 16.7 & 17.3 & 43.7 \\
    Hoodie   & 35.7 & 32.3 & 31.0 & 29.5 & 19.8 & 20.1 & 18.9 & 35.9 \\
    \midrule
    Average  & \textbf{47.2} & \textbf{45.8} & 36.6 & 38.0 & \textbf{15.6} & 16.7 & \textbf{15.4} & 31.4 \\
    \bottomrule
    \end{tabular}
    \vspaceafter
\end{table*}

\subsubsection{Experiment Setup}

\textbf{\\}\vspace{-3.5mm}

\textbf{Environment.} In this experiment, the model is given an image and the garment category and is required to predict the positions of all keypoints. We use PyFlex~\cite{li2018learning} as the physics simulator and Blender~\cite{blender} for rendering. By synthesizing garment images and keypoint annotations in simulation, we train a model to predict keypoints and then directly evaluate it on real-world images without any fine-tuning on real data. We assume that the mask is known and the background is masked out. For real-world images, we use Grounded-SAM~\cite{ren2024grounded} for segmentation. Models trained on synthetic datasets are directly tested on this real-world dataset without fine-tuning. 

\textbf{Asset.} For each garment category, we generate 1,500 synthetic garment instances for training using our proposed asset generation pipeline. We manually collected and annotated 480 t-shirts, 82 trousers, 96 vests, and 96 hoodies to construct a real-world test dataset. 

\textbf{Metric.} We select \textbf{Mean Average Precision (meanAP)} and \textbf{Average Keypoint Distance (AKD)} as metrics~\cite{lips2024learningkeypointsroboticcloth}. We model keypoint detection as a classification problem, where a keypoint is considered correctly classified if the pixel-wise Euclidean distance between the detected keypoint and the ground truth is below a given threshold. The thresholds in our experiments are 4 pixels and 8 pixels, which correspond to approximately \SI{0.5}{\centi\metre} and \SI{1}{\centi\metre} in the real world, respectively. Under this definition, meanAP is the proportion of keypoints that are correctly classified, and the reported meanAP is the average over the two thresholds. AKD is the average pixel-wise distance between predicted and ground truth keypoints. In the ground truth images, we only annotate visible keypoints, and for both metrics, invisible keypoints are ignored during evaluation. 

\textbf{Data generation cost.} In our data generation pipeline, creating a mesh template is very fast. Deforming a mesh using PyFlex takes approximately 6 seconds, and generating a texture image with Stable-Diffusion-3.5 requires around 20 seconds (excluding the time waiting for ChatGPT responses). Rendering a 480 × 720 image takes about 2 seconds. Overall, it takes roughly 30 seconds to generate a single cloth instance. All experiments are conducted on a server equipped with two Intel Xeon Platinum 8255C CPUs (48 cores, 96 threads, 2.50 GHz base frequency) and an NVIDIA RTX 3090 GPU. 

\textbf{Training details.} The keypoint coordinates are converted into Gaussian blobs on 2D probability heatmaps, which serve as targets for pixel-wise logistic regression using a binary cross-entropy loss. We directly adopt the model architecture from~\cite{lips2024learningkeypointsroboticcloth}, which is a U-Net~\cite{ronneberger2015unetconvolutionalnetworksbiomedical}-inspired architecture, with a pretrained MaxViT~\cite{tu2022maxvitmultiaxisvisiontransformer} nano model as the encoder. The network takes a $480\times720$ masked RGB image as input and outputs $N$ heatmaps of size $256\times256$, where $N$ denotes the number of keypoints for the garment category. Keypoints are extracted from the predicted heatmaps by identifying local maxima within a $3\times3$ pixel window, using a probability threshold of 0.01. During training, we apply data augmentation techniques including color jittering, random rotation and translation, and random patching. The model is trained for 5 hours per category on a single NVIDIA RTX 3090 GPU. 

\begin{figure}[h]
    \centering
    \includegraphics[width=1.0\linewidth]{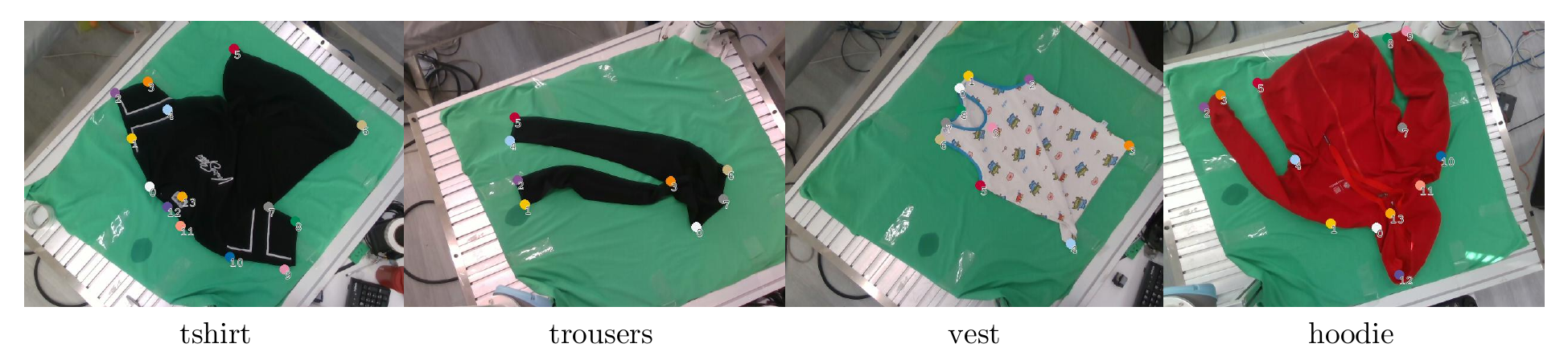}
    \caption{\textbf{Qualitative results of keypoint detection on real images.} The figure shows the predicted output of our keypoint detection model on real images.}
    \label{fig:keypoint_detection_real}
    \vspaceafter
\end{figure}

\subsubsection{Experiment Results}

\textbf{\\}\vspace{-3.5mm}

In this experiment, we address the following two questions: 

\textbf{How do the textures generated by our method compare with those produced by other approaches?} We study this question by training the same model on datasets generated using different garment texture methods: \textit{Ours}, \textit{aRTF}~\cite{lips2024learningkeypointsroboticcloth}, and \textit{Paint-it}~\cite{youwang2024paintit}. 

\textbf{Can filtering with a VLM improve the appearance quality of the generated meshes?} We include results from our pipeline both with and without the final filtering stage. 

\begin{figure}[h]
    \centering
    \includegraphics[width=1.0\linewidth]{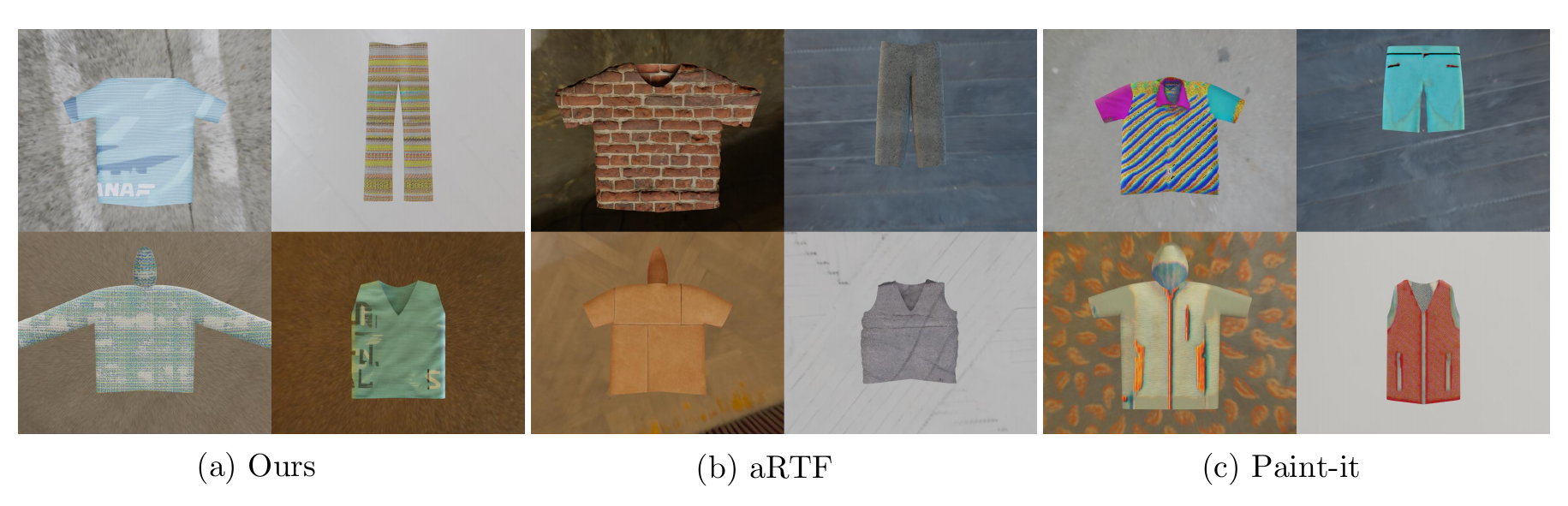}
    \caption{\textbf{Examples of generated meshes.} Compared with other texture generation methods, our approach produces textures that are generally more plausible.}
    \label{fig:asset_compare}
    \vspaceafter
\end{figure}

The experimental results are presented in Table~\ref{tab:keypoint_detection_performance}, and some keypoint detection results on real images are shown in Figure~\ref{fig:keypoint_detection_real}. The results indicate that, compared with other approaches, our framework produces garments with more realistic appearances and achieves strong performance on both the meanAP and AKD metrics. The ablation study regarding the final filtering stage further demonstrates the effectiveness of the VLM-based filtering step. Some qualitative comparisons of the generated meshes are shown in Figure~\ref{fig:asset_compare}.  

\subsection{Folding Policy Learning}

\subsubsection{Experiment Setup}

\textbf{\\}\vspace{-3.5mm}

\textbf{Environment.} We use PyFlex~\cite{li2018learning} as the physics simulator and Blender~\cite{blender} for rendering. The initial garment state includes random rotation around the z-axis (vertical axis), random flipping, and randomly generated wrinkles. We use the RGB images from the robot's head-mounted D436 camera as single-view visual input. The complete action space consists of the XYZ coordinates of both grippers, as well as the grippers’ open–close states, resulting in a total of 8 dimensions. We use inverse kinematics (IK) to compute the robot’s joint angles from the end-effector pose. During IK solving, we constrain only the grippers to remain parallel to the table, leaving the other two rotational degrees of freedom unconstrained. The table height is assumed to be known and fixed. Each demonstration trajectory has approximately 120 steps. During testing, a trajectory is considered terminated if it exceeds 300 steps or if the movement distance between consecutive actions is less than \SI{1}{\milli\meter}. 

\begin{figure}[h]
    \centering
    \includegraphics[width=1.0\linewidth]{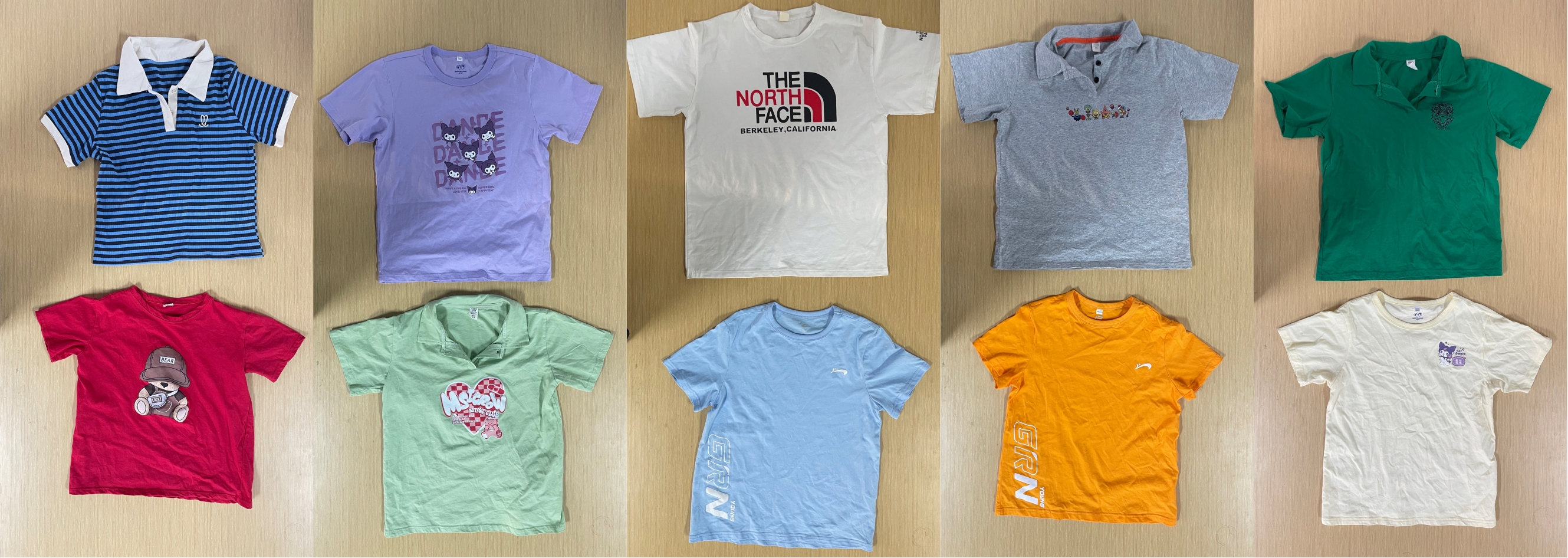}
    \caption{\textbf{Real-world assets for garment folding.} In the real-world experiments, each garment is folded twice, and the average success rate is computed. }
    \label{fig:real_world_assets}
    \vspaceafter
\end{figure}

\textbf{Asset.} For each garment category, we generate 1,000 training instances using our proposed asset generation pipeline. During testing, an additional set of 100 previously unseen garments is used. For the table and scene backgrounds, we randomly sample a collection of indoor assets downloaded from PolyHaven~\cite{polyhaven}. For real-world testing, we use 10 unseen garments, as shown in Figure~\ref{fig:real_world_assets}. 

\begin{figure*}[t]
    \centering
    \includegraphics[width=1.0\linewidth]{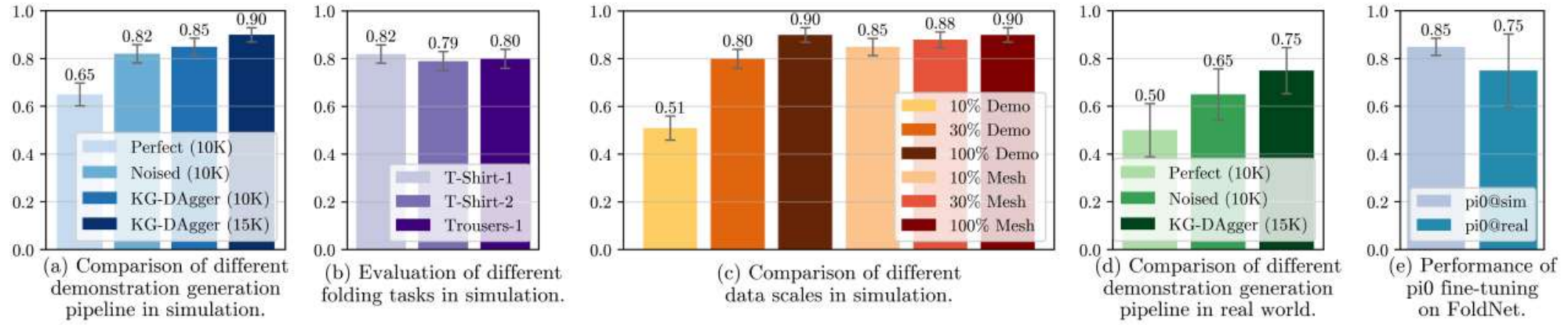}
    \caption{\textbf{Quantitative results of garment folding.} We compare the average success rates of garment folding tasks for different models in simulation and in the real world.}
    \label{fig:exp_tab}
    \vspaceafter
\end{figure*}

\textbf{Metric.} To automatically determine folding success in simulation, we first run the keypoint-based policy on a perfectly initialized garment configuration. The resulting garment mesh is referred to as \textit{mesh\_gt}. During model evaluation, the final folded mesh, \textit{mesh\_eval}, is compared with \textit{mesh\_gt}. After aligning the two meshes by an arbitrary rigid-body rotation and translation, we compute the Euclidean distances between all pairs of corresponding vertices and define their average as the evaluation metric. Folding is considered successful if the average vertex distance is below \SI{0.4}{\milli\meter}. In real-world experiments, folding success is determined by human experts.

\textbf{Data generation cost.} Simulation and rendering are computationally intensive for this task, with high demands on both CPU and GPU resources. We use AMD EPYC 7543 CPUs (128 cores in total) and 8 NVIDIA RTX 4090 GPUs to perform the simulation and rendering. Generating 1,000 trajectories requires approximately one day. 

\textbf{Training Details.} We use a CNN-based policy from Diffusion Policy~\cite{chi2023diffusionpolicy}. The observation encoder employs ResNet50, producing observation features with a dimensionality of 512. Simultaneously, the robot's current state (the XYZ coordinates of the left and right end-effectors and the grippers’ open–close status) is mapped to a 512-dimensional space. The observation and state features are then concatenated to form the conditional input to the diffusion policy. The model uses only the current observation and proprioception as input. Its output is an action sequence of length 16, and during inference, we execute the first four actions. Depending on the dataset size and the specific task, the total number of training steps ranges from approximately 100k to 400k. Training requires approximately one day on 8 NVIDIA RTX 4090 GPUs. 

\subsubsection{Experiment Results}

\textbf{\\}\vspace{-3.5mm}

In this experiment, we address the following five questions: 

\textbf{How does the proposed KG-DAgger improve the quality of training data?} We evaluate this by employing different methods for generating demonstrations and comparing the performance of models trained with the same amount of data. 

\textbf{Within our data generation framework, can new folding rules be designed to enable the model to learn alternative garment folding strategies?} We devise different folding strategies and evaluate the success rate of each. 

\textbf{What is the trend of success rate with respect to the amount of training data?} We compare model performance under varying amounts of training data and training meshes. 

\textbf{Can the model transfer to the real world?} We evaluate the success rates of models trained with different data generation methods in real-world scenarios. 

\textbf{Can the VLA model be fine-tuned with FoldNet?} We load the pretrained $\pi_0$~\cite{pi0} model, a mainstream large VLA model, and fine-tune it on the FoldNet dataset. We evaluate the model on a robot unseen by the original $\pi_0$ model, conducting tests in both simulation and the real world.

\begin{figure}[h]
    \centering
    \includegraphics[width=1.0\linewidth]{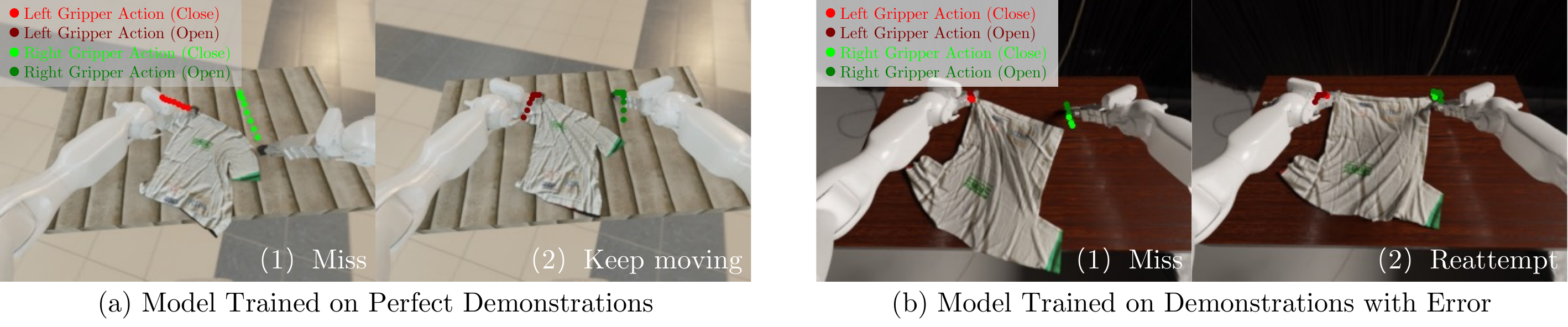}
    \caption{\textbf{Comparison of models trained with different demonstration data generation methods in simulation.} In (a), the training data do not include recovery strategies for errors, so a failed grasp results in an out-of-distribution situation. In (b), the training data include recovery strategies, allowing the model to retry grasping after a failure.}
    \label{fig:sim_fold_result}
    \vspaceafter
\end{figure}

\paragraph{Different demonstration generation pipelines} In Figure~\ref{fig:exp_tab}(a), we compare the performance of models trained with different demonstration generation pipelines in simulation. \textit{Perfect} refers to using only perfect demonstrations, while \textit{Noised} refers to demonstrations generated by adding noise to ground-truth actions~\cite{lyuscissorbot}. This baseline also employs the keypoint-based error recovery strategy to augment the dataset but differs from KG-DAgger in that the actions are obtained by perturbing the ground-truth actions before execution, rather than using actions predicted by the network. \textit{KG-DAgger} corresponds to the complete method described in Section~\ref{sec:demonstration}. The numbers in parentheses indicate the total number of training trajectories used. 

When the dataset includes trajectories with error corrections (\textit{Noised}, \textit{KG-DAgger}), there is a significant performance improvement compared to using only perfect demonstrations (\textit{Perfect}). Figure~\ref{fig:sim_fold_result} illustrates this difference: the model trained exclusively on perfect demonstrations fails to retry after a grasp failure, whereas the model trained with error recovery data succeeds. Moreover, the \textit{KG-DAgger} method further reduces the gap between the training and testing data distributions compared to the \textit{Noised} method, leading to better performance. 

\paragraph{Different tasks} Our data generation pipeline can be adapted to various folding patterns. In Figure~\ref{fig:exp_tab}(b), we use 10K trajectories generated by the \textit{Noised} method as training data. The folding procedures for \textit{T-Shirt-1} and \textit{Trousers-1} are shown in Figure~\ref{fig:keypoint_based_policy}(a). The difference between \textit{T-Shirt-2} and \textit{T-Shirt-1} lies in the final step: instead of folding the bottom of the shirt upward, \textit{T-Shirt-2} folds it from left to right. The experimental results show that our framework is not limited to a specific folding method. 

\paragraph{Data scale} Figure~\ref{fig:exp_tab}(c) illustrates how the model’s performance varies with the amount of training data. Here, 100\% usage indicates training with 1000 garments and 15K demonstrations. When varying the number of meshes or demonstrations, the quantity of the other is kept fixed. 

\begin{figure}[h]
    \centering
    \includegraphics[width=1.0\linewidth]{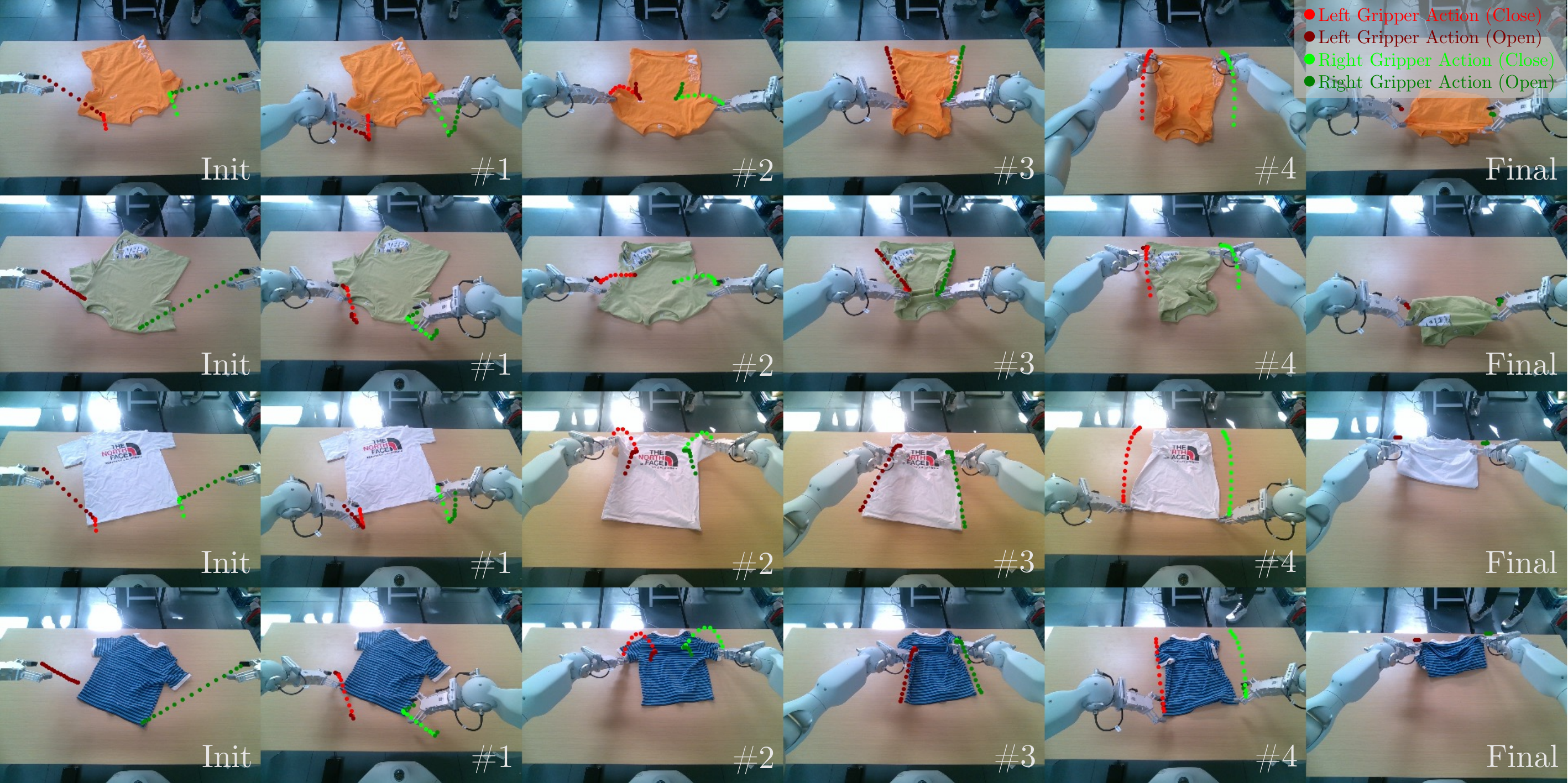}
    \caption{\textbf{Real-world deployment.} The figure illustrates the performance of our policy in real-world scenarios.}
    \label{fig:real_fold_result}
    \vspaceafter
\end{figure}

\paragraph{Sim2real performance} Our trained model can be directly transferred from simulation to the real world. As shown in Figure~\ref{fig:exp_tab}(d), we compare the real-world performance of models trained with different demonstration generation methods. The model trained with our KG-DAgger approach outperforms those trained with other demonstration data. Representative examples of model outputs in real-world experiments are shown in Figure~\ref{fig:real_fold_result}. 

\paragraph{Fine-tuning VLA with FoldNet} We directly fine-tuned the $\pi_0$ model on our dataset, with the language input fixed as \textit{"Fold the T-shirt."}. The pre-trained $\pi_0$ model we used has 3 billion parameters and employs a flow matching head. The rest of the model’s inputs and outputs are consistent with those of the DP model. We use a batch size of 64 and a learning rate of 2.5e-5, fine-tuning all parameters for 50,000 steps, which takes approximately 16 hours on 8 H100 GPUs. The experimental results are shown in Figure~\ref{fig:exp_tab}(e). The results demonstrate that even without using any real-world data, we can still train a VLA model capable of generalizing to real-world scenarios.


\section{Conclusions}
\label{sec:conclusion}

In this paper, we present a synthetic dataset for garment folding. At the core of the dataset are garment keypoints, which enable both the synthesis of garment meshes and the generation of demonstration data. To further improve model performance, we incorporate keypoint-based error recovery data into the demonstration dataset. Our experiments show that models trained with this dataset can be directly transferred to real robots and unseen garments.


\section{Limitation}

\begin{figure}[h]
    \centering
    \includegraphics[width=1.0\linewidth]{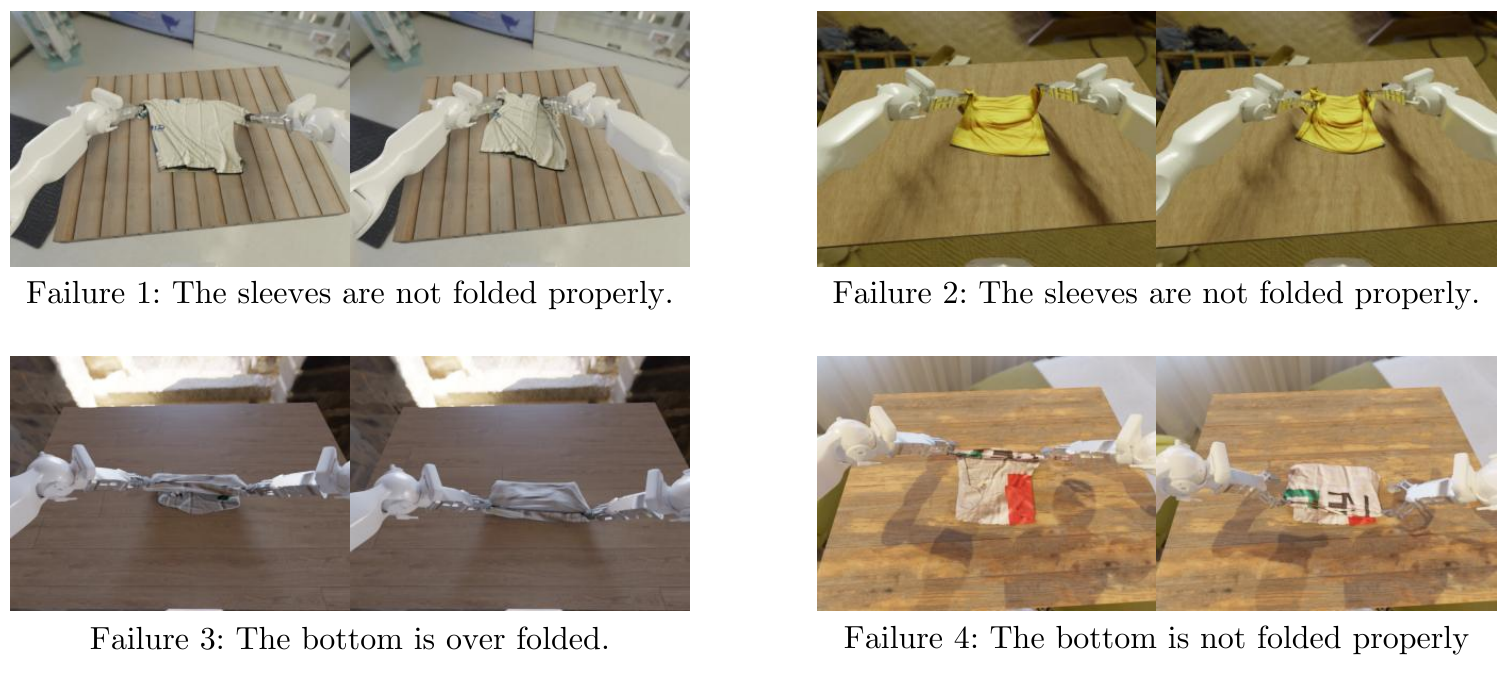}
    \caption{Failure mode in simulation.}
    \label{fig:failure_mode_sim}
    \vspaceafter
\end{figure}

\begin{figure}[h]
    \centering
    \includegraphics[width=1.0\linewidth]{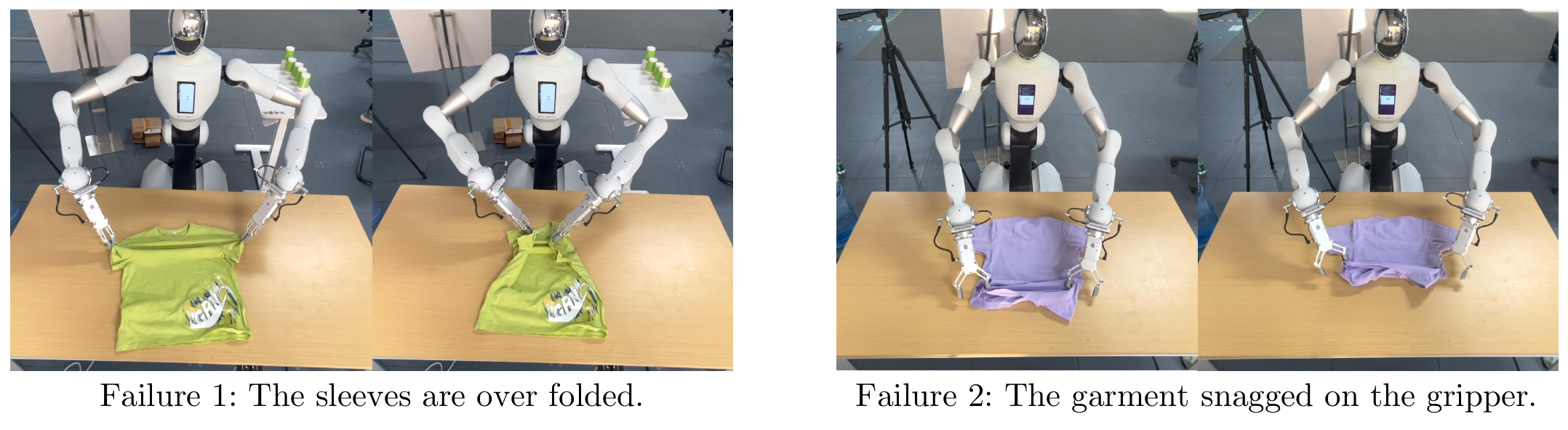}
    \caption{Failure mode in real world.}
    \label{fig:failure_mode_real}
    \vspaceafter
\end{figure}

Although KG-DAgger improves the model’s ability to recover from failures, certain failure modes remain challenging. Representative examples are shown in Figure~\ref{fig:failure_mode_sim} and Figure~\ref{fig:failure_mode_real}. In particular, some unexpected situations in the real world are difficult to accurately reproduce in simulation.

Currently, the folding patterns are relatively simple, mainly due to limitations in the physical realism of the cloth simulation. When more complex folding methods are adopted, the realism of the simulation significantly degrades. In the future, the use of finer cloth meshes and more efficient and accurate simulators could further reduce the sim-to-real gap. Exploring the addition of rotational degrees of freedom in the action space is also a promising direction. In addition, combining synthetic and real-world data has the potential to improve success rates in the real world.


\bibliographystyle{bibtex/bst/IEEEtran}
\bibliography{bibtex/bib/IEEEabrv, bibtex/bib/all}

@inproceedings{chi2023diffusionpolicy,
	title={Diffusion Policy: Visuomotor Policy Learning via Action Diffusion},
	author={Chi, Cheng and Feng, Siyuan and Du, Yilun and Xu, Zhenjia and Cousineau, Eric and Burchfiel, Benjamin and Song, Shuran},
	booktitle={Proceedings of Robotics: Science and Systems (RSS)},
	year={2023}
}

@inproceedings{li2018learning,
  author       = {Yunzhu Li and
                  Jiajun Wu and
                  Russ Tedrake and
                  Joshua B. Tenenbaum and
                  Antonio Torralba},
  title        = {Learning Particle Dynamics for Manipulating Rigid Bodies, Deformable
                  Objects, and Fluids},
  booktitle    = {7th International Conference on Learning Representations, {ICLR} 2019,
                  New Orleans, LA, USA, May 6-9, 2019},
  year         = {2019},
}

@misc{blender,
  author       = {{Blender Foundation}},
  title        = {Blender},
  url          = {https://www.blender.org},
}

@misc{polyhaven,
  author       = {{PolyHaven Team}},
  title        = {Poly Haven - The Public 3D Asset Library},
  url          = {https://polyhaven.com/},
}

@inproceedings{lu2024garmentlab,
 author = {Lu, Haoran and Wu, Ruihai and Li, Yitong and Li, Sijie and Zhu, Ziyu and Ning, Chuanruo and Shen, Yan and Luo, Longzan and Chen, Yuanpei and Dong, Hao},
 booktitle = {Advances in Neural Information Processing Systems},
 pages = {11866--11903},
 publisher = {Curran Associates, Inc.},
 title = {GarmentLab: A Unified Simulation and Benchmark for Garment Manipulation},
 OPTurl = {https://proceedings.neurips.cc/paper_files/paper/2024/file/15f80ec0fed53885d2ca6272edb96ede-Paper-Conference.pdf},
 volume = {37},
 year = {2024},
 OPTeditor = {A. Globerson and L. Mackey and D. Belgrave and A. Fan and U. Paquet and J. Tomczak and C. Zhang},
}

@article{chen2024deformpam,
  author       = {Wendi Chen and
                  Han Xue and
                  Fangyuan Zhou and
                  Yuan Fang and
                  Cewu Lu},
  title        = {DeformPAM: Data-Efficient Learning for Long-horizon Deformable Object
                  Manipulation via Preference-based Action Alignment},
  journal      = {CoRR},
  volume       = {abs/2410.11584},
  year         = {2024},
  OPTurl          = {https://doi.org/10.48550/arXiv.2410.11584},
  doi          = {10.48550/ARXIV.2410.11584},
  eprinttype    = {arXiv},
  eprint       = {2410.11584},
  OPTbiburl       = {https://dblp.org/rec/journals/corr/abs-2410-11584.bib},
  bibsource    = {dblp computer science bibliography, https://dblp.org}
}

@inproceedings{lyuscissorbot,
  title     = {ScissorBot: Learning Generalizable Scissor Skill for Paper Cutting via Simulation, Imitation, and Sim2Real},
  author    = {Lyu, Jiangran and Chen, Yuxing and Du, Tao and Zhu, Feng and Liu, Huiquan and Wang, Yizhou and Wang, He},
  booktitle = {Proceedings of The 8th Conference on Robot Learning},
  year      = {2024}
}

@inproceedings{xue2023unifoldingsampleefficientscalablegeneralizable,
  title     = {UniFolding: Towards Sample-efficient, Scalable, and Generalizable Robotic Garment Folding},
  author    = {Xue, Han and Li, Yutong and Xu, Wenqiang and Li, Huanyu and Zheng, Dongzhe and Lu, Cewu},
  booktitle = {Proceedings of The 7th Conference on Robot Learning},
  year      = {2023}
}

@INPROCEEDINGS{wang2023policydressalllearning,
    AUTHOR    = {Yufei Wang AND Zhanyi Sun AND Zackory Erickson AND David Held}, 
    TITLE     = {{One Policy to Dress Them All: Learning to Dress People with Diverse Poses and Garments}}, 
    BOOKTITLE = {Proceedings of Robotics: Science and Systems}, 
    YEAR      = {2023}, 
    ADDRESS   = {Daegu, Republic of Korea}, 
    MONTH     = {July}, 
    DOI       = {10.15607/RSS.2023.XIX.008} 
}

@INPROCEEDINGS{avigal2022speedfoldinglearningefficientbimanual,
  author={Avigal, Yahav and Berscheid, Lars and Asfour, Tamim and Kröger, Torsten and Goldberg, Ken},
  booktitle={2022 IEEE/RSJ International Conference on Intelligent Robots and Systems (IROS)}, 
  title={SpeedFolding: Learning Efficient Bimanual Folding of Garments}, 
  year={2022},
  volume={},
  number={},
  pages={1-8},
  doi={10.1109/IROS47612.2022.9981402}
}

@ARTICLE{zare2024surveyimitation,
  author={Zare, Maryam and Kebria, Parham M. and Khosravi, Abbas and Nahavandi, Saeid},
  journal={IEEE Transactions on Cybernetics}, 
  title={A Survey of Imitation Learning: Algorithms, Recent Developments, and Challenges}, 
  year={2024},
  volume={54},
  number={12},
  pages={7173-7186},
  doi={10.1109/TCYB.2024.3395626}}

@article{longhini2024unfoldingliteraturereviewrobotic,
  author       = {Alberta Longhini and
                  Yufei Wang and
                  Irene Garcia{-}Camacho and
                  David Blanco{-}Mulero and
                  Marco Moletta and
                  Michael C. Welle and
                  Guillem Aleny{\`{a}} and
                  Hang Yin and
                  Zackory Erickson and
                  David Held and
                  J{\'{u}}lia Borr{\`{a}}s and
                  Danica Kragic},
  title        = {Unfolding the Literature: {A} Review of Robotic Cloth Manipulation},
  journal      = {Annu. Rev. Control. Robotics Auton. Syst.},
  volume       = {8},
  number       = {1},
  pages        = {295--322},
  year         = {2025},
  OPTurl          = {https://doi.org/10.1146/annurev-control-022723-033252},
  doi          = {10.1146/ANNUREV-CONTROL-022723-033252},
  OPTbiburl       = {https://dblp.org/rec/journals/arcras/LonghiniWGBMWAYEHBK25.bib},
  bibsource    = {dblp computer science bibliography, https://dblp.org}
}

@ARTICLE{lips2024learningkeypointsroboticcloth,
  author={Lips, Thomas and De Gusseme, Victor-Louis and Wyffels, Francis},
  journal={IEEE Robotics and Automation Letters}, 
  title={Learning Keypoints for Robotic Cloth Manipulation Using Synthetic Data}, 
  year={2024},
  volume={9},
  number={7},
  pages={6528-6535},
  doi={10.1109/LRA.2024.3405335}
}

@INPROCEEDINGS{canberk2022clothfunnels,
  author={Canberk, Alper and Chi, Cheng and Ha, Huy and Burchfiel, Benjamin and Cousineau, Eric and Feng, Siyuan and Song, Shuran},
  booktitle={2023 IEEE International Conference on Robotics and Automation (ICRA)}, 
  title={Cloth Funnels: Canonicalized-Alignment for Multi-Purpose Garment Manipulation}, 
  year={2023},
  volume={},
  number={},
  pages={5872-5879},
  doi={10.1109/ICRA48891.2023.10161546}
}

@article{tian2025diffusiondynamicsmodelsgenerative,
  author       = {Tongxuan Tian and
                  Haoyang Li and
                  Bo Ai and
                  Xiaodi Yuan and
                  Zhiao Huang and
                  Hao Su},
  title        = {Diffusion Dynamics Models with Generative State Estimation for Cloth
                  Manipulation},
  journal      = {CoRR},
  volume       = {abs/2503.11999},
  year         = {2025},
  OPTurl          = {https://doi.org/10.48550/arXiv.2503.11999},
  doi          = {10.48550/ARXIV.2503.11999},
  eprinttype    = {arXiv},
  eprint       = {2503.11999},
  OPTbiburl       = {https://dblp.org/rec/journals/corr/abs-2503-11999.bib},
  bibsource    = {dblp computer science bibliography, https://dblp.org}
}

@misc{jiang2025phystwinphysicsinformedreconstructionsimulation,
      title={PhysTwin: Physics-Informed Reconstruction and Simulation of Deformable Objects from Videos}, 
      author={Hanxiao Jiang and Hao-Yu Hsu and Kaifeng Zhang and Hsin-Ni Yu and Shenlong Wang and Yunzhu Li},
      year={2025},
      eprint={2503.17973},
      archivePrefix={arXiv},
      primaryClass={cs.CV},
      OPTurl={https://arxiv.org/abs/2503.17973}, 
}

@INPROCEEDINGS{zhou2023clothesnetinformationrich3dgarment,
  author={Zhou, Bingyang and Zhou, Haoyu and Liang, Tianhai and Yu, Qiaojun and Zhao, Siheng and Zeng, Yuwei and Lv, Jun and Luo, Siyuan and Wang, Qiancai and Yu, Xinyuan and Chen, Haonan and Lu, Cewu and Shao, Lin},
  booktitle={2023 IEEE/CVF International Conference on Computer Vision (ICCV)}, 
  title={ClothesNet: An Information-Rich 3D Garment Model Repository with Simulated Clothes Environment}, 
  year={2023},
  volume={},
  number={},
  pages={20371-20381},
  doi={10.1109/ICCV51070.2023.01868}}

@inproceedings{bertiche2020cloth3d,
author = {Bertiche, Hugo and Madadi, Meysam and Escalera, Sergio},
title = {CLOTH3D: Clothed 3D Humans},
year = {2020},
isbn = {978-3-030-58564-8},
publisher = {Springer-Verlag},
address = {Berlin, Heidelberg},
OPTurl = {https://doi.org/10.1007/978-3-030-58565-5_21},
doi = {10.1007/978-3-030-58565-5_21},
booktitle = {Computer Vision – ECCV 2020: 16th European Conference, Glasgow, UK, August 23–28, 2020, Proceedings, Part XX},
pages = {344–359},
numpages = {16},
location = {Glasgow, United Kingdom}
}

@inproceedings{youwang2024paintit,
    title = {Paint-it: Text-to-Texture Synthesis via Deep Convolutional Texture Map Optimization and Physically-Based Rendering},
    author = {Youwang, Kim and Oh, Tae-Hyun and Pons-Moll, Gerard},
    booktitle = {IEEE Conference on Computer Vision and Pattern Recognition (CVPR)},
    year = {2024}
}

@article{luo2024hilserl,
    author = {Jianlan Luo  and Charles Xu  and Jeffrey Wu  and Sergey Levine },
    title = {Precise and dexterous robotic manipulation via human-in-the-loop reinforcement learning},
    journal = {Science Robotics},
    volume = {10},
    number = {105},
    pages = {eads5033},
    year = {2025},
    doi = {10.1126/scirobotics.ads5033},
    OPTurl = {https://www.science.org/doi/abs/10.1126/scirobotics.ads5033},
    eprint = {https://www.science.org/doi/pdf/10.1126/scirobotics.ads5033},
}

@inproceedings{hejna2024remixoptimizingdatamixtures,
  title     = {ReMix: Optimizing Data Mixtures for Large Scale Imitation Learning},
  author    = {Hejna, Joey and Bhateja, Chethan Anand and Jiang, Yichen and Pertsch, Karl and Sadigh, Dorsa},
  booktitle = {Proceedings of The 8th Conference on Robot Learning},
  year      = {2024}
}

@article{myers2024policyadaptationlanguageoptimization,
  author       = {Vivek Myers and
                  Bill Chunyuan Zheng and
                  Oier Mees and
                  Sergey Levine and
                  Kuan Fang},
  title        = {Policy Adaptation via Language Optimization: Decomposing Tasks for
                  Few-Shot Imitation},
  journal      = {CoRR},
  volume       = {abs/2408.16228},
  year         = {2024},
  OPTurl          = {https://doi.org/10.48550/arXiv.2408.16228},
  doi          = {10.48550/ARXIV.2408.16228},
  eprinttype    = {arXiv},
  eprint       = {2408.16228},
  OPTbiburl       = {https://dblp.org/rec/journals/corr/abs-2408-16228.bib},
  bibsource    = {dblp computer science bibliography, https://dblp.org}
}

@INPROCEEDINGS{kelly2019hgdaggerinteractiveimitationlearning,
  author={Kelly, Michael and Sidrane, Chelsea and Driggs-Campbell, Katherine and Kochenderfer, Mykel J.},
  booktitle={2019 International Conference on Robotics and Automation (ICRA)}, 
  title={HG-DAgger: Interactive Imitation Learning with Human Experts}, 
  year={2019},
  volume={},
  number={},
  pages={8077-8083},
  doi={10.1109/ICRA.2019.8793698}
}

@article{ren2024grounded,
  author       = {Tianhe Ren and
                  Shilong Liu and
                  Ailing Zeng and
                  Jing Lin and
                  Kunchang Li and
                  He Cao and
                  Jiayu Chen and
                  Xinyu Huang and
                  Yukang Chen and
                  Feng Yan and
                  Zhaoyang Zeng and
                  Hao Zhang and
                  Feng Li and
                  Jie Yang and
                  Hongyang Li and
                  Qing Jiang and
                  Lei Zhang},
  title        = {Grounded {SAM:} Assembling Open-World Models for Diverse Visual Tasks},
  journal      = {CoRR},
  volume       = {abs/2401.14159},
  year         = {2024},
  OPTurl          = {https://doi.org/10.48550/arXiv.2401.14159},
  doi          = {10.48550/ARXIV.2401.14159},
  eprinttype    = {arXiv},
  eprint       = {2401.14159},
  OPTbiburl       = {https://dblp.org/rec/journals/corr/abs-2401-14159.bib},
  bibsource    = {dblp computer science bibliography, https://dblp.org}
}

@inproceedings{podell2023sdxlimprovinglatentdiffusion,
  author       = {Dustin Podell and
                  Zion English and
                  Kyle Lacey and
                  Andreas Blattmann and
                  Tim Dockhorn and
                  Jonas M{\"{u}}ller and
                  Joe Penna and
                  Robin Rombach},
  title        = {{SDXL:} Improving Latent Diffusion Models for High-Resolution Image
                  Synthesis},
  booktitle    = {The Twelfth International Conference on Learning Representations,
                  {ICLR} 2024, Vienna, Austria, May 7-11, 2024},
  year         = {2024},
}

@misc{openai2024gpt4ocard,
      title={GPT-4o System Card}, 
      author={OpenAI},
      year={2024},
      eprint={2410.21276},
      archivePrefix={arXiv},
      primaryClass={cs.CL},
      url={https://arxiv.org/abs/2410.21276}, 
}

@INPROCEEDINGS{wen2024foundationposeunified6dpose,
  author={Wen, Bowen and Yang, Wei and Kautz, Jan and Birchfield, Stan},
  booktitle={2024 IEEE/CVF Conference on Computer Vision and Pattern Recognition (CVPR)}, 
  title={FoundationPose: Unified 6D Pose Estimation and Tracking of Novel Objects}, 
  year={2024},
  volume={},
  number={},
  pages={17868-17879},
  doi={10.1109/CVPR52733.2024.01692}
}

@InProceedings{ronneberger2015unetconvolutionalnetworksbiomedical,
    author="Ronneberger, Olaf
    and Fischer, Philipp
    and Brox, Thomas",
    title="U-Net: Convolutional Networks for Biomedical Image Segmentation",
    booktitle="Medical Image Computing and Computer-Assisted Intervention -- MICCAI 2015",
    year="2015",
    publisher="Springer International Publishing",
    address="Cham",
    pages="234--241",
    OPTeditor="Navab, Nassir
    # and Hornegger, Joachim
    # and Wells, William M.
    # and Frangi, Alejandro F.",
}

@article{tu2022maxvitmultiaxisvisiontransformer,
  title={MaxViT: Multi-Axis Vision Transformer},
  author={Tu, Zhengzhong and Talebi, Hossein and Zhang, Han and Yang, Feng and Milanfar, Peyman and Bovik, Alan and Li, Yinxiao},
  journal={ECCV},
  year={2022},
}

@misc{pi0,
      title={$\pi_0$: A Vision-Language-Action Flow Model for General Robot Control}, 
      author={{Physical Intelligence Team}},
      year={2024},
      eprint={2410.24164},
      archivePrefix={arXiv},
      primaryClass={cs.LG},
      url={https://arxiv.org/abs/2410.24164}, 
}

@article{liu2024rdt,
    title={RDT-1B: a Diffusion Foundation Model for Bimanual Manipulation},
    author={Liu, Songming and Wu, Lingxuan and Li, Bangguo and Tan, Hengkai and Chen, Huayu and Wang, Zhengyi and Xu, Ke and Su, Hang and Zhu, Jun},
    journal={arXiv preprint arXiv:2410.07864},
    year={2024}
}

@misc{kim2024openvlaopensourcevisionlanguageactionmodel,
      title={OpenVLA: An Open-Source Vision-Language-Action Model}, 
      author={{OpenVLA Model Team}},
      year={2024},
      eprint={2406.09246},
      archivePrefix={arXiv},
      primaryClass={cs.RO},
      url={https://arxiv.org/abs/2406.09246}, 
}

@inproceedings{octomodelteam2024octoopensourcegeneralistrobot,
    title={Octo: An Open-Source Generalist Robot Policy},
    author = {{Octo Model Team}},
    booktitle = {Proceedings of Robotics: Science and Systems},
    address  = {Delft, Netherlands},
    year = {2024},
}

@article{objaverseXL,
  title={Objaverse-XL: A Universe of 10M+ 3D Objects},
  author={Matt Deitke and Ruoshi Liu and Matthew Wallingford and Huong Ngo and
          Oscar Michel and Aditya Kusupati and Alan Fan and Christian Laforte and
          Vikram Voleti and Samir Yitzhak Gadre and Eli VanderBilt and
          Aniruddha Kembhavi and Carl Vondrick and Georgia Gkioxari and
          Kiana Ehsani and Ludwig Schmidt and Ali Farhadi},
  journal={arXiv preprint arXiv:2307.05663},
  year={2023}
}

@INPROCEEDINGS{chen2025metafoldlanguageguidedmulticategorygarment,
  author={Haonan Chen and Junxiao Li and Ruihai Wu and Yiwei Liu and Yiwen Hou and Zhixuan Xu and Jingxiang Guo and Chongkai Gao and Zhenyu Wei and Shensi Xu and Jiaqi Huang and Lin Shao},
  booktitle={2025 IEEE/RSJ International Conference on Intelligent Robots and Systems (IROS)}, 
  title={MetaFold: Language-Guided Multi-Category Garment Folding Framework via Trajectory Generation and Foundation Model}, 
  year={2025},
}
\end{document}